# Memristor Crossbar-based Hardware Implementation of Fuzzy Membership Functions

Farnood Merrikh-Bayat
Electrical Engineering Department
Sharif University of Technology
Tehran, Iran
F_merrikhbayat@ee.sharif.edu

Saeed Bagheri Shouraki
Electrical Engineering Department
Sharif University of Technology
Tehran, Iran
Bagheri-s@sharif.edu

*Abstract*—In May 1, 2008, researchers at Hewlett Packard (HP) announced the first physical realization of a fundamental circuit element called memristor that attracted so much interest worldwide. This newly found element can easily be combined with crossbar interconnect technology which this new structure has opened a new field in designing configurable or programmable electronic systems. These systems in return can have applications in signal processing and artificial intelligence. In this paper, based on the simple memristor crossbar structure, we propose new and simple circuits for hardware implementation of fuzzy membership functions. In our proposed circuits, these fuzzy membership functions can have any shapes and resolutions. In addition, these circuits can be used as a basis in the construction of evolutionary systems.

## I. INTRODUCTION

Nowadays, fuzzy inference systems are extensively in use in so many different applications [1, 2] and for variety of reasons such as modeling and control [3]. The main advantage of these systems compared to other conventional methods is their capacity to tolerate information expressed in a way that is uncertain and imprecise.

Almost all of the currently working fuzzy systems are implemented digitally on digital hardware devices such as Field-Programmable Gate Arrays (FPGAs) [4], Digital Signal Processors (DSPs) and dedicated Application Specific Integrated Circuits (ASICs) [5]. However, such computing paradigms suffer from the constant need of establishing a trade-off between flexibility and performance. In these kinds of devices, the arithmetic operations are carried out with limited computational precision [6–8]. Another drawback of these digital hardware devices is that they are very poor in acting as an evolvable hardware. In addition, they are completely in contrast with the nature of the fuzzy which is uncertainty. In this paper, we propose a new and simple hardware based on the memristor crossbar structure for implementing fuzzy membership functions. Since any operation is performed in analog in our proposed circuits, computational precision is theoretically infinite. Here, it is worth to mention that since the fuzzy membership functions are implemented through the memristance of the memristors in the crossbar, they can be simply evolved by changing their memristance value.

The paper is organized as follows. In Section 2, we describe the characteristics of memristors and their application in the field of programmable analog circuits. Section 3 is devoted to the description of our proposed memristor crossbar-based hardware implementation of fuzzy membership functions. Finally, a brief conclusion is presented in Section 4.

## II. MEMRISTORS AND THEIR CONTRIBUTION IN CONSTRUCTING PROGRAMMABLE ANALOG CIRCUITS

Memristor is an electrically switchable semiconductor thin film sandwiched between two metal contacts with a total length of $D$ and consists of doped and un-doped regions which its physical structure with its equivalent circuit model is shown in Fig. 1 [9]. The internal state variable $W$ determines the length of doped region with low resistance against un-doped region with high resistivity. This internal state variable and consequently the total resistivity of the device can be changed by applying external voltage bias $V(t)$. This means that passing current from memristor in one direction will increase the resistance while changing the direction of the applied current will decrease its memristance [10]. On the other hand, it is obvious that in this element, passing current in one direction for longer period of time will change the resistance of the memristor more.

As a result, memristor is nothing else than the analog variable resistor which its resistance can be adjusted by changing the direction and duration of the applied voltage or current. Therefore, memristor can be used as a storage device

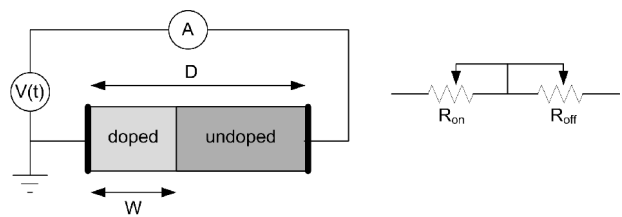

Figure 1. Structure of the memristor reported by HP researchers and its equivalent circuit model.

in which analog values can be stored as impedance instead of voltage.

These analog memories can be simply combined by the crossbar structure. A crossbar array basically consists of two sets of conductive parallel wires intersecting each other perpendicularly. The intersections (or crosspoints) are separated by a thin film material which its properties such as its resistance can be changed by controlling the voltage applied to it such as memristor. Now, it is well-known that these memristor crossbar structures can offer so many new applications in the field of programmable electronic systems [11, 12].

### III. MEMRISTOR CROSSBAR-BASED HARDWARE IMPLEMENTATION OF FUZZY MEMBERSHIP FUNCTIONS

In this section, we will show that how any membership function with any shape can be implemented with a simple memristor crossbar. Assume that $A$ is a fuzzy set defined on the input domain $x$ and $\mu_A(\cdot)$ denotes the membership function of it. In fact, by hardware implementation of membership function we mean that we are going to construct a system that if the input sample $x_0$ is presented to this system, the output becomes $\mu_A(x_0)$. Hereafter, we will illustrate the idea of using memristor crossbar for implementing membership functions by some examples.

First of all, consider a simplest case in which two fuzzy sets $A$ and $B$ with membership functions $\mu_A(\cdot)$ and $\mu_B(\cdot)$ respectively are defined on input domain $x$ as shown in Fig. 2(a). In almost all of the current applications, input signal is sampled and quantized with finite resolution. Without loss of generality, in this example we assume that the input is quantized with a quantizer which its quantization step is equal to 1. In this case, the input variable $x$ can only have discrete integer values and therefore, fuzzy sets $A$ and $B$ and their corresponding membership functions $\mu_A(\cdot)$ and $\mu_B(\cdot)$ will become discrete as well as shown in Fig. 2(b) for convenience. Figure 2(c) shows the memristor crossbar-based circuit that we propose as a hardware implementation of fuzzy sets of Fig. 2(a). In this figure, the crossbar has two rows where each of them corresponds to one of the fuzzy sets $A$ and $B$. Fuzzy membership function $\mu_A(\cdot)$ is implemented in lower row and fuzzy membership function $\mu_B(\cdot)$ is implemented in upper row of this crossbar. In addition, each column of this crossbar represents a discrete value of the input variable $x$. Memristors at those crosspoints which are specified by black dots should be programmed to have memristance values equal to those which are shown in Fig. 2(c) (written near to them). In fact, memristance of the memristor located at the crossing point of the lower horizontal wire and the vertical wire corresponds to $x_0$ should be equal to $R/\mu_A(x_0)$ where $\mu_A(x_0)$ is the membership function of the fuzzy set $A$ at point $x_0$ and $R$ is the feedback resistor of the opamps of Fig. 2(c). Any other memristors in this crossbar which is not programmed ($\mu_A(x_0)$ is zero) will have the memristance of $R_{off}$ where $R_{off}$ is the highest possible memristance value of the memristor.

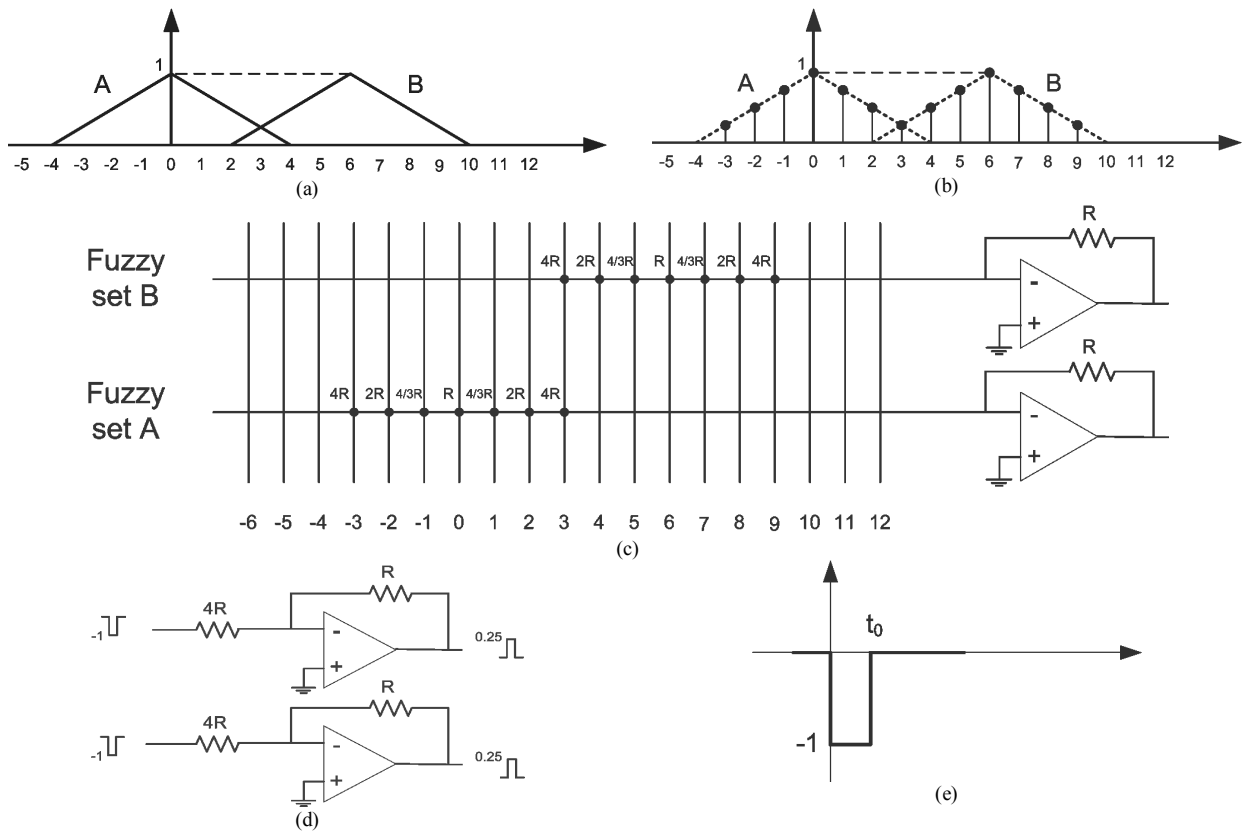

Figure 2. First example of memristor crossbar-based hardware implementation of membership functions; (a) Continuous form of the membership functions; (b) Discrete form of the membership functions; (c) Memristor crossbar-based hardware implementation of membership functions; (d) Equivalent circuit; (e) A negative narrow pulse with amplitude -1 which is applied to the columns of the crossbar.

The circuit of Fig. 2(c) works as follows. Suppose that the input sample $x_0 = 3$ is observed. In this case, a negative narrow pulse with amplitude -1 such as the one shown in Fig. 2(e) is applied to the $10^{th}$ column of the crossbar (which corresponds to $x = 3$) while other columns are connected to high impedance. Note that the width of the pulse should be chosen in a way that its application does not change the memristance of the memristors. In this case, the memristor crossbar of Fig. 2(c) will be equal to the circuit shown in Fig. 2(d) which in this configuration, outputs of opamps $A$ and $B$ are $0.25$ and $0.25$ respectively. It is evident that these outputs are equal to $\mu_A(x_0 = 3)$ and $\mu_B(x_0 = 3)$. Therefore, in the circuit of Fig. 2(c), the input is the quantized sample of the input variable $x$ and the outputs are the membership functions of fuzzy sets $A$ and $B$, i.e. $\mu_A(x)$ and $\mu_B(x)$. Now, consider a situation in which the input observed sample is $x_0 = 8$. In this case, the output of the lower opamp will be $R/R_{off}$ whereas ideally it should be zero. Consequently, resistance $R$ should be chosen in a way that $R/R_{off}$ be as much small as possible. It is worth to mention that in this example, samples of the input variable are considered to be a singleton instead of the fuzzy number (the reason of applying negative narrow pulse to one column while connecting others to high impedance).

Figure 3 shows another example of hardware implementation of fuzzy membership functions. Figure 3(a) shows the continuous form of the fuzzy sets and their corresponding membership functions while the discrete versions of them are depicted in Fig. 3(b). Similar to the previous example, these fuzzy sets simply can be implemented by our proposed structure like the one demonstrated in Fig. 3(c). By applying a negative pulse of Fig. 2(e) to any column of this circuit, values of the membership functions at that point will appear at the outputs of the opamps.

Through these two examples, it should become clear that any number of fuzzy sets with any shapes can be simply implemented by our memristor crossbar-based circuits. In addition, it is evident that by increasing the number of columns of the crossbar, it is possible to increase the resolution of the input signal. For instance, assume that we use finer quantizer for the quantization of the input signal, i.e. $x$ in our first example. In this case, if the quantization step of the utilized quantizer becomes $\Delta q = 0.5$, then the two fuzzy sets in Fig. 2 can be implemented such as the one shown in Fig. 4.

In all of the above examples, the input signal is assumed to be a singleton. Now, consider the case in which the input signal is a fuzzy number. In this case, the output of the structure should be a fuzzy number as well. As an example, consider the typical fuzzy set and the input fuzzy number which are shown in Fig. 5(a) and Fig. 5(b) respectively. In its simplest form, at the output we should have two fuzzy numbers which are the componentwise multiplication of the input fuzzy number with the membership functions of two fuzzy sets. This process can simply be implemented by the memristor crossbar-based circuit shown in Fig. 5(c). Memristance of the memristors at those crosspoints which are specified by black dots in Fig. 5(c) should be adjusted to the values written near to them in this figure. Any other memristors in the crossbar should have their highest memristance value, *i.e.* $R_{off}$. In fact by this way, we implement the fuzzy set of Fig. 5(a) in the antidiagonal of the crossbar through the memristance of the memristor. Now, if the samples of the fuzzy number of Fig. 5(b) are interpreted as a voltage and are applied to the columns of the crossbar, voltages at the outputs of the opamps will be the output fuzzy numbers. By applying standard opamp circuit analysis techniques, it is easy to see that the outputs of the opamps are the samples of the componentwise multiplication of the fuzzy sets (Fig. 5(a)) and fuzzy input number (Fig. 5(b)).

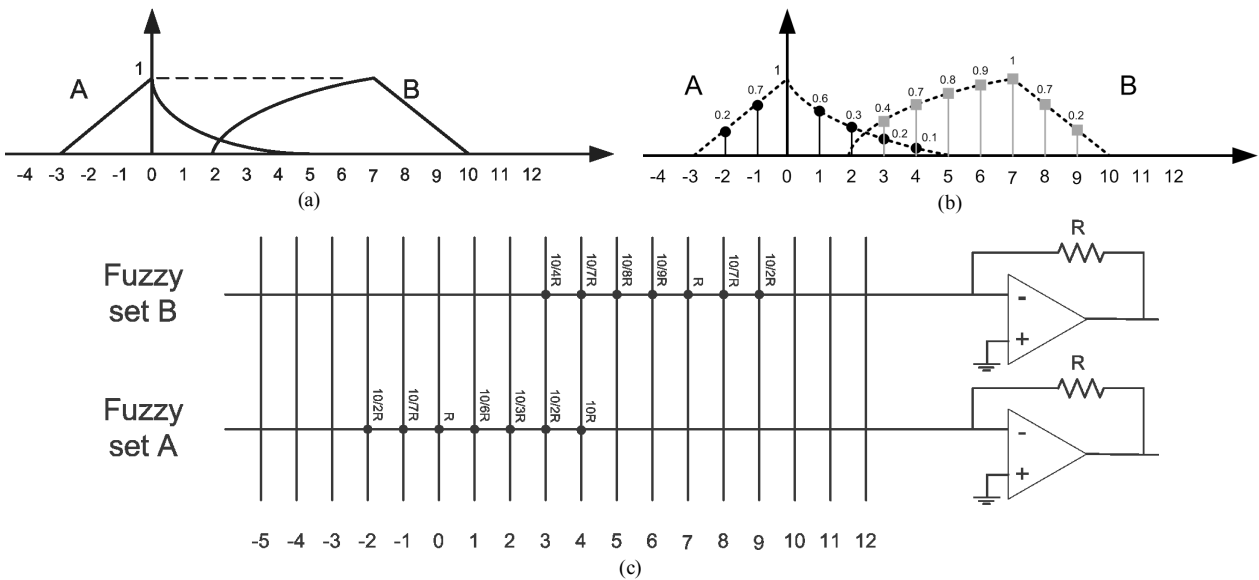

Figure 3. Another example demonstrating how any number of membership functions with any shapes can be implemented by memristor crossbars.

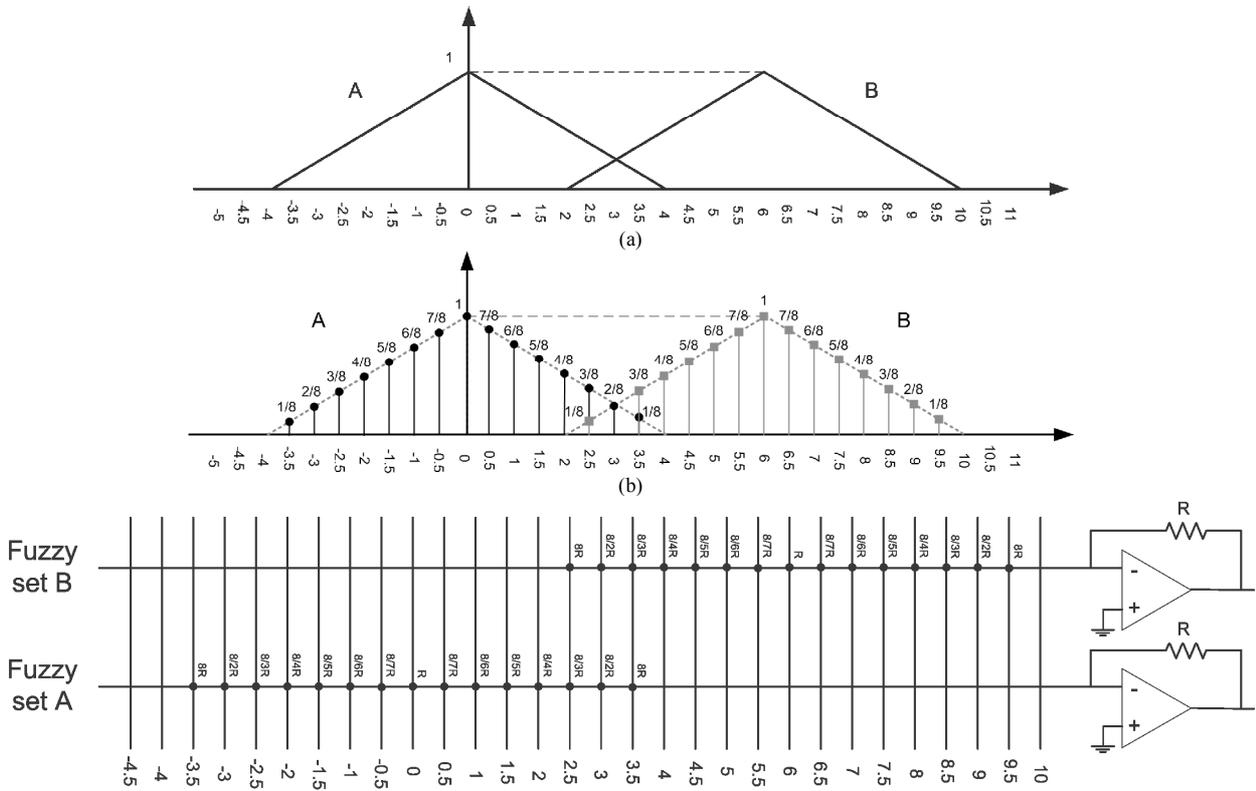

Figure 4. Hardware implementation of membership functions of Fig. 2(a) with higher resolution.

Here it is worth to mention that memristor crossbar structures like the one proposed in this paper have this potential that they can be used as a platform for implementing evolvable hardware. This is because of the fact that in these kinds of systems, variable parameters are mostly implemented at the crosspoints through the memristance of the memristors which can in return be simply modified by applying the suitable voltage or current. Note that modification of the memristance of the memristors can be done even during the execution time of the system.

## IV. CONCLUSION

In this paper we illustrated how any number of membership functions with any shapes and resolution can simply be implemented through the memristor crossbar-based structure. In our proposed method, membership functions are programmed into the memristance of the memristors at the crosspoints. One of the most particular advantages of our proposed circuits is that they have a capability of being used in evolutionary systems.

REFERENCES

[1] M. Hanmandlu, O. P. verma, N. K. kumar and M. kulkarni, "A novel optimal fuzzy system for color image enhancement using bacterial foraging," *IEEE Trans. On Instrumentation and Measurment*, vol. 58, No. 8, pp. 2867–2879, 2009.

[2] J. pan, G. N. DeSouza, and A. C. Kak, "fuzzyshell: a large-scale expert system shell using fuzzy logic for uncertainty reasoning," *IEEE Trans. On Fuzzy Syst.*, vol. 6, pp. 563–581, Nov. 1998.

[3] C. C. Lee, "Fuzzy logic in control systems: Fuzzy Logic Controller-(Parts 1 and 2)," *IEEE Trans. On Syst., man, Cyber.*, vol. 20, pp. 404–432, 1990.

[4] D. Kim, "An implementation of fuzzy logic controller on the reconfigurable FPGA system," *IEEE Trans. On Industrial Electronics*, vol. 47, No. 3, pp. 703–715, 2002.

[5] D. L. Hung, "dedicated digital fuzzy hardware," *IEEE Micro*, vol. 15, No. 4, pp. 31–39, 1995.

[6] J.M. Cioffi, "Limited-precision effects in adaptive filtering," IEEE Transactions on circuits and systems, CAS-34(7):821833, 1987.

[7] S. Haykin, "Adaptive filter theory," Prentice-Hall, second edition, 1991.

[8] J.R. Treichler, C.R. Johnson, and M.G. Larimore, "Theory and design of adaptive filters," Wiley Interscience, New York, 1987.

[9] D.B. Strukov, G.S. Snider, D.R. Stewart and R.S. Williams, "The missing memristor found," Nature, 2008, vol. 453, pp. 80–83, 1 May 2008.

[10] L.O. Chua, "Memristor - the missing circuit element," IEEE Trans. on Circuit Theory, vol. CT-18, no. 5, pp. 507–519, 1971.
[11] F. Merrikh-Bayat, and S. Bagheri Shouraki, "Mixed analog-digital crossbar-based hardware implementation of sign–sign LMS adaptive filter," Analog Integrated Circuits and Signal Processing, DOI 10.1007/s10470-010-9523-3.
[12] S. Shin, K. Kim, and S.M. Kang, "Memristor-based fine resolution resistance and its applications," ICCCAS 2009, July 2009.

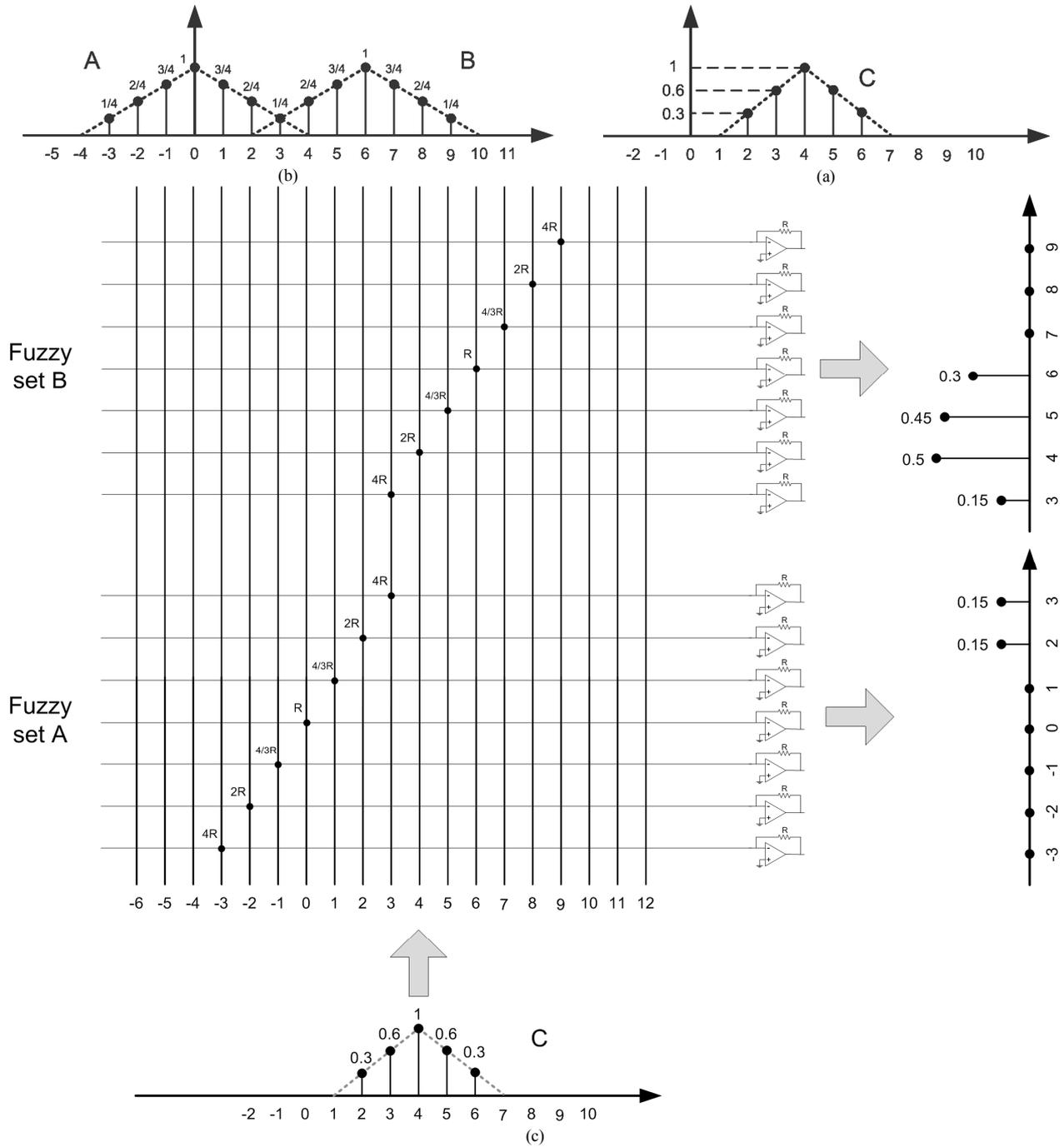

Figure 5. Memristor crossbar-based hardware implementation of membership functions in the case that both of the input and output of the system are fuzzy numbers.